\documentclass[a4paper,twoside]{article}

\usepackage{epsfig}
\usepackage{subfigure,ctable}
\usepackage{calc}
\usepackage{amssymb}
\usepackage{amstext}
\usepackage{amsmath}
\usepackage{amsthm}
\usepackage{multicol,multirow}
\usepackage{pslatex}
\usepackage{apalike}
\usepackage{epstopdf}
\usepackage{SCITEPRESS}     

\usepackage{hyperref}
\usepackage[]{graphicx}
\usepackage{enumerate} 

\begin{document}

\title{Domain Specific Author Attribution Based on Feedforward Neural Network Language Models}
\author{\authorname{Zhenhao Ge and Yufang Sun}
\affiliation{School of Electrical and Computer Engineering, Purdue University, West Lafayette, Indiana, U.S.A}
\email{zhenhao.ge@gmail.com, sun361@purdue.edu}
}

\keywords{Authorship Attribution, Neural Networks, Language Modeling}

\abstract{Authorship attribution refers to the task of automatically determining the author based on a given sample of text. It is a problem with a long history and has a wide range of application. Building author profiles using language models is one of the most successful methods to automate this task. New language modeling methods based on neural networks alleviate the curse of dimensionality and usually outperform conventional N-gram methods. However, there have not been much research applying them to authorship attribution. In this paper, we present a novel setup of a Neural Network Language Model (NNLM) and apply it to a database of text samples from different authors. We investigate how the NNLM performs on a task with moderate author set size and relatively limited training and test data, and how the topics of the text samples affect the accuracy. NNLM achieves nearly 2.5\% reduction in perplexity, a measurement of fitness of a trained language model to the test data. Given 5 random test sentences, it also increases the author classification accuracy by 3.43\% on average, compared with the N-gram methods using SRILM tools.  An open source implementation of our methodology is freely available at \url{https://github.com/zge/authorship-attribution/}.}

\onecolumn \maketitle \normalsize \vfill

\section{\uppercase{Introduction}}
\label{sec:introduction}

\noindent Authorship attribution refers to the task of identifying the text author from a given text sample, by finding the author's unique textual features. It is possible to do this because the author's profile or style embodies many characteristics, including personality, cultural and educational background, language origin, life experience and knowledge basis, etc. Every person has his/her own style, and sometimes the author's identity can be easily recognized. However, most often identifying the author is challenging, because author's style can vary significantly by topics, mood, environment and experience. Seeking consistency or consistent evolution out of variation is not always an easy task.  

There has been much research in this area. Juola \cite{juola2006authorship} and Stamatatos \cite{stamatatos2009survey} for example, have surveyed the state of the art and proposed a set of recommendations to move forward. As more text data become available from the Web and computational linguistic models using statistical methods mature, more opportunities and challenges arise in this area \cite{koppel2009computational}. Many statistical models have been successfully applied in this area, such as Latent Dirichlet Allocation (LDA) for topic modeling and dimension reduction  \cite{seroussi2011authorship}, Naive Bayes for text classification \cite{coyotl2006authorship},  Multiple Discriminant Analysis (MDA) and Support Vector Machines (SVM) for feature selection and classification \cite{ebrahimpour2013automated}. Methods based on language modeling are also among the most popular methods for authorship attribution \cite{kevselj2003n}.

Neural networks with deep learning have been successfully applied in many applications, such as speech recognition \cite{hinton2012deep}, object detection \cite{krizhevsky2012imagenet}, natural language processing \cite{socher2011parsing}, and other pattern recognition and classification tasks \cite{bishop1995neural}, \cite{ge2015sleep}. Neural Network based Language Models (NNLM) have surpassed the performance of traditional N-gram LMs \cite{bengio2003neural}, \cite{mnih2007three} and are purported to generalize better in smaller datasets \cite{mnih2010learning}. In this paper, we propose a similar NNLM setup for authorship attribution. The performance of the proposed method depends highly on the settings of the experiment, in particular the experimental design, author set size and data size \cite{luyckx2011scalability}. In this work, we focused on small datasets within one specific text domain, where the sizes of the training and test datasets for each author are limited. This often leads to context-biased models, where the accuracy of author detection is highly dependent on the degree to which the topics in training and test sets match each other \cite{luyckx2008authorship}. The experiments we conceive are based on a closed dataset, i.e. each test author also appears in the training set, so the task is simplified to author classification rather than detection. 

The paper is organized as follows. Sec. \ref{sec:data} introduces the database used for this project. Sec. \ref{sec:algorithm} explains the methodology of the NNLM, including cost function definition, forward-backward propagation, and weight and bias updates. Sec. \ref{sec:results} describes the implementation of the NNLM, provides the classification metrics, and compares results with conventional baseline N-gram models. Finally, Sec. \ref{sec:conclusion} presents the conclusion and suggests future work.

\section{\uppercase{Data Preparation}}
\label{sec:data}

\noindent The database is a selection of course transcripts from Coursera, one of the largest Massive Open Online Course (MOOC) platforms. To ensure the author detection less replying on the domain information, 16 courses were selected from one specific text domain of the technical science and engineering fields, covering 8 areas: Algorithm, Data Mining, Information Technologies (IT), Machine Learning, Mathematics, Natural Language Processing (NLP), Programming and Digital Signal Processing (DSP). Table \ref{tab:data_profile} lists more details for each course in the database, such as the number of sentences and words, the number of words per sentence, and vocabulary sizes in multiple stages. For privacy reason, the exact course titles and instructor (author) names are concealed. However, for the purpose of detecting the authors, it is necessary to point out that all courses are taught by different instructors, except for the courses with IDs 7 and 16. This was done intentionally to allow us to investigate how the topic variation affects performance. 

\begin{table}[!htb]
\centering
\caption{Subtitle database from selected Coursera courses}
\label{tab:data_profile}
\setlength{\tabcolsep}{2pt}
\scriptsize
\renewcommand{\arraystretch}{1}
\begin{tabular}{@{} *{4}{l}cc @{}} \toprule%
\multirow{2}*{ID} & \multirow{2}*{Field} & No. of    & No. of & Words / &  Vocab. size (original \\
   &       & sentences & words  & sentences  &  / stemmed / pruned) \\\midrule
1 & Algorithm & 5,672 & 121,675 & 21.45 & 3,972 / 2,702 / 1,809 \\
2 & Algorithm & 14,902 & 294055 & 20.87 & 6,431 / 4,222 / 2,378 \\
3 & DSP & 8,126 & 129,665 & 15.96 & 3,815 / 2,699 / 1,869 \\
4 & Data Mining & 7,392	& 129,552 & 17.53 & 4,531 / 3,140 / 2,141 \\
5 & Data Mining & 6,906 & 129,068 & 18.69 & 3,008 / 2,041 / 1,475 \\
6 & DSP & 20,271 & 360,508 & 17.78 & 8,878 / 5,820 / 2,687 \\
7 & IT & 9,103 & 164,812 & 18.11 & 4,369 / 2,749 / 1,979 \\
8 & Mathematics & 5,736 & 101,012 & 17.61 & 3,095 / 2,148 / 1,500 \\
9 & Machine Learning & 11,090 & 224,504 & 20.24 & 6,293 / 4,071 / 2,259 \\ 
10 & Programming & 8,185 & 160,390 & 19.60 & 4,045 / 2,771 / 1,898 \\
11 & NLP & 7,095 & 111,154 & 15.67 & 3,691 / 2,572 / 1,789 \\
12 & NLP & 4,395 & 100,408 & 22.85 & 3,973 / 2,605 / 1,789 \\
13 & NLP & 4,382 & 96,948 & 22.12 & 4,730 / 3,467 / 2,071 \\
14 & Machine Learning & 6,174 & 116,344 & 18.84 & 5,844	/ 4,127 / 2,686 \\
15 & Mathematics & 5,895 & 152,100 & 25.80 & 3,933 / 2,697 / 1,918 \\
16 & Programming & 6,400 & 136,549 & 21.34 & 4,997 / 3,322 / 2,243 \\\bottomrule
\end{tabular}
\end{table}

\begin{figure}[!htb]
\centering
\includegraphics[width=0.48\textwidth]{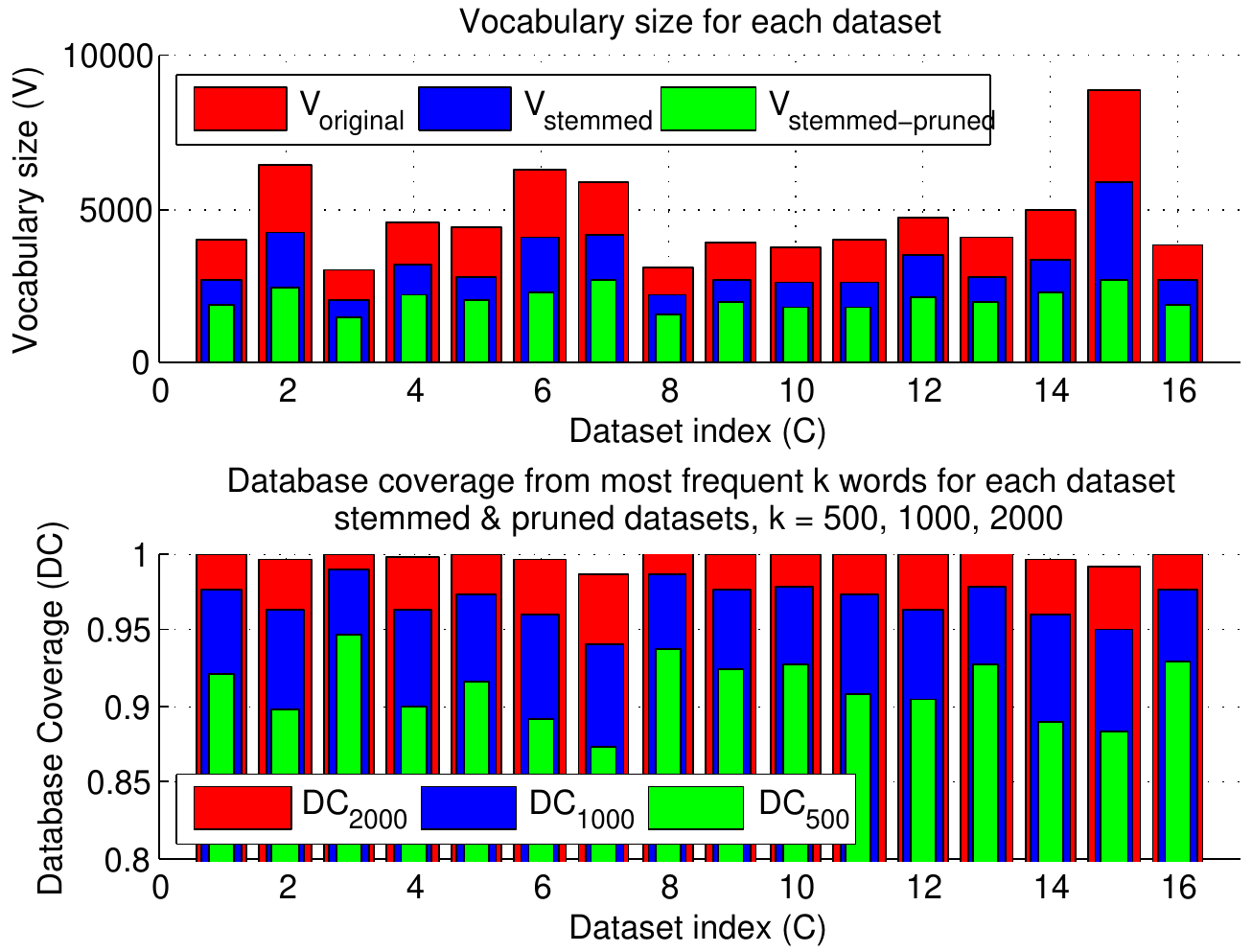}
\caption{Database profile with respect to vocabulary size and word coverage in various stages}
\label{fig:data_profile}
\end{figure}

The transcripts for each course were originally collected in short phrases with various lengths, shown one at a time at the bottom of the video lectures. They were first concatenated and then segmented into sentences, using straight-forward boundary determination by punctuations. The sentence-wise datasets are then stemmed using the  Porter Stemming algorithm \cite{porter1980algorithm}. To further control the vocabulary size, words occurring only once in the entire course or with frequency less than $1/100,000$ are considered to have negligible influence on the outcome and are pruned by mapping them to an Out-Of-Vocabulary (OOV) mark $\langle{\textrm{unk}}\rangle$. The first top bar graph in Figure \ref{fig:data_profile} shows how the vocabulary size of each course dataset shrinks after stemming and pruning. There are only $0.5\sim1.5\%$ words among all datasets mapped to $\langle{\textrm{unk}}\rangle$, however, the vocabulary sizes are significantly reduced to an average of $2000$. The bottom bar graph provides a profile of each instructor in terms of word frequency, i.e. the database coverage of the most frequent $k$ words after stemming and pruning, where $k = 500, 1000, 2000$. For example, the most frequent 500 words cover at least 85\% of the words in all datasets. 

\section{\uppercase{Neural Network Language Model}}
\label{sec:algorithm}

\begin{figure}[htb]
\centering
\includegraphics[height=5.5cm, width=7.5cm]{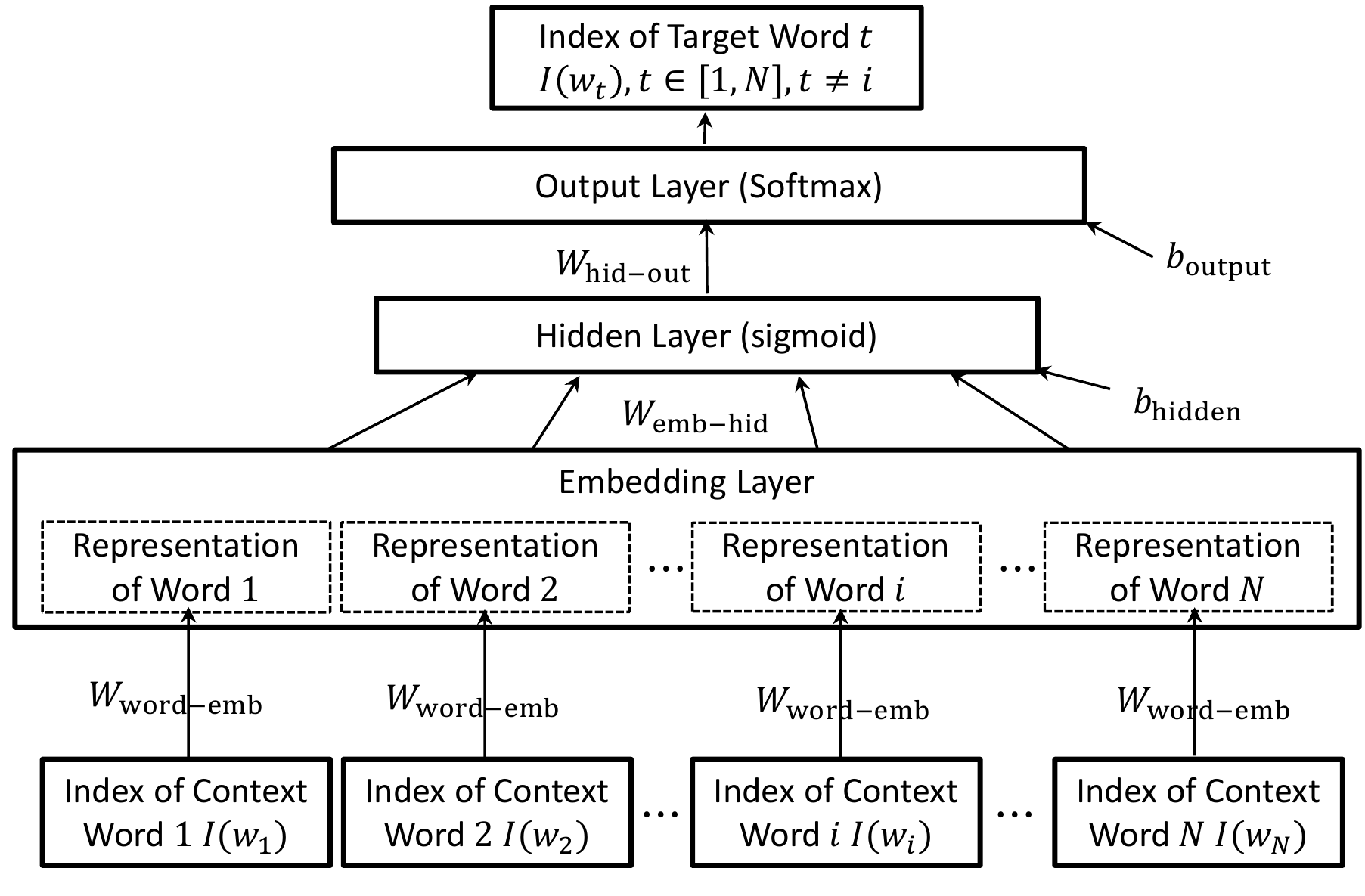}
\caption{Architecture of the Neural Network Language Model ($I$: index, $\mathcal{W}$: word, $N$: number of context words, $W$: weight, $b$: bias)}
\label{fig:nnlm}
\end{figure}

\noindent The language model is trained using a feed-forward neural network illustrated in Figure \ref{fig:nnlm}. Given a sequence of $N$ words $\mathcal{W}_{1}, \mathcal{W}_{2}, \ldots, \mathcal{W}_{i}, \ldots,  \mathcal{W}_{N}$ from training text, the network trains weights to predict the word $\mathcal{W}_t$, $t \in [1, N]$ in a designated target word position  in sequence, using the information provided from the rest of words, as it is formulated in Eq. (\ref{eq:concept}). 
\begin{eqnarray}
\label{eq:concept}
\mathcal{W}^{*} = \arg\max_{t} P(\mathcal{W}_{t}|\mathcal{W}_{1}\mathcal{W}_{2} \cdots \mathcal{W}_{i} \cdots \mathcal{W}_{N}), \: i \neq t
\end{eqnarray}
It is similar to the classic N-gram language model, where the primary task is to predict the next word given $N-1$ previous words. However, here the network can be trained to predict the target word in any position, given the neighboring words. 

The network contains 4 different types of layers: the word layer, the embedding layer, the hidden layer, and the output (softmax) layer. The weights between adjacent layers, i.e. word-to-embedding weights $W_\mathrm{word-emb}$, embedding-to-hidden weights $W_\mathrm{emb-hid}$, and hidden-to-output weights $W_\mathrm{hid-out}$, need to be trained in order to transform the input words to the predicted output word. The following 3 sub-sections briefly introduce the NNLM training procedure, first defining the cost function to be minimized, then describing the forward and backward weight and bias propagation. The implementation details regarding  parameter settings and tuning are discussed in Sec. \ref{sec:results}.

\subsection{Cost Function}

\noindent Given vocabulary size $V$, it is a multinomial classification problem to predict a single word out of $V$ options. So the cost function to be minimized can be formulated as
\begin{equation}
\label{eq:costfunction}
C = -\sum_{V} t_{j} \log y_{j} .
\end{equation}
$C$ is the cross-entropy, and $y_{j}$, where $j \in V$ and $\sum_{j \in V} y_{j} = 1$, is the output of node $j$ in the final output layer of the network, i.e. the probability of selecting the $j$th word as the predicted word. The parameter $t_{j}$ is the target label and $t_{j} \in \{0,1\}$. As a 1-of-V multi-class classification problem, there is only one target value $1$, and the rest are $0$s. 

\subsection{Forward Propagation}

Forward propagation is a process to compute the outputs $y_{j}$ of each layer $L_{j}$ with a) its neural function (i.e. sigmoid, linear, rectified, binary, etc.), and b) the inputs $z_{j}$, computed using the outputs of the previous layer $y_{i}$, weights $W_{ij}$ from layer $L_{i}$ to layer $L_{j}$, and bias $b_{j}$ of the current layer $L_{j}$. After weight and bias initialization, the neural network training starts from forward propagating the word inputs to the outputs in the final layer. 

For the word layer, given word context size $N$ and target word position $t$, each of the $N-1$ input words $w_{i}$ is represented by a binary index column vector $x_{i}$ with length equal to the vocabulary size $V$. It contains all $0$s but only one $1$ in a particular position to differentiate it from all other words. The word $x_{i}$ is transformed to its distributed representation in the so-called embedding layer via the equation

\begin{equation}
\label{eq:emb_input}
z_\mathrm{emb}(i) = W_\mathrm{word-emb}^{T} \cdot x_{i} , 
\end{equation}
where $W_\mathrm{word-emb}$ is the word-to-embedding weights with size $[V \times N_\mathrm{emb}]$, which is used in the computation of $z_\mathrm{emb}(i)$ for different words $x_{i}$, and $N_\mathrm{emb}$ is the dimension of the embedding space. Because $z_\mathrm{emb}(i)$ is one column in $W_\mathrm{word-emb}^{T}$, representing the word $x_{i}$, this process is simply a table look up.

For the embedding layer, the output $y_\mathrm{emb}$ is just the concatenation of the representation of the input words $z_\mathrm{emb}(i)$,  
\begin{equation}
\label{eq:emb_state}
y_\mathrm{emb} = [z_\mathrm{emb}^{T}(1), z_\mathrm{emb}^{T}(2), \cdots, 
   z_\mathrm{emb}^{T}(i), \cdots, z_\mathrm{emb}^{T}(N)]^{T} ,
\end{equation}
%
%
where $i \in V$, $i \neq t$, and $t$ is the index for the target word $w_{t}$. So $y_\mathrm{emb}$ is a column vector with length $N_\mathrm{emb} \times  (N-1)$.

For the hidden layer, the input $z_\mathrm{hid}$ is firstly computed with weights $W_\mathrm{emb-hid}$, embedding output $y_\mathrm{emb}$, and hidden bias $b_\mathrm{hid}$ using
\begin{equation}
\label{eq:hid_input}
z_\mathrm{hid} = W_\mathrm{emb-hid}^{T} \cdot y_\mathrm{emb} + b_\mathrm{hid} ,
\end{equation}
Then, the logistic function, which is a type of Sigmoid function, is used to compute the output $y_\mathrm{hid}$ from $z_\mathrm{hid}$:
\begin{equation}
\label{eq:hid_output}
y_\mathrm{hid} = \frac{1}{1 + e ^ {-z_\mathrm{hid}}} . 
\end{equation}

For the output layer, the input $z_\mathrm{out}$ is given by
\begin{equation}
\label{eq:out_input}
z_\mathrm{out} = W_\mathrm{hid-out}^{T} \cdot y_\mathrm{hid} + b_\mathrm{out},
\end{equation} 
This output layer is a Softmax layer which incorporates the constraint $\sum_{V} y_\mathrm{out} = 1$ using the Softmax function
\begin{equation}
\label{eq:softmax}
y_\mathrm{out} = \frac{e^{z_\mathrm{out}}}{\sum_{V} e^ {z_\mathrm{out}}} .
\end{equation}

\subsection{Backward Propagation}
After forward propagating the input words $x_{i}$ to the final output $y_\mathrm{out}$ of the network, through Eq. (\ref{eq:emb_input}) to Eq. (\ref{eq:softmax}), the next task is to backward propagate error derivatives from the output layer to the input, so that we know the directions and magnitudes to update weights between layers. 

It starts from the derivative $\frac{\partial C}{\partial z_\mathrm{out}(i)}$ of node $i$ in the output layer, i.e. 
\begin{equation}
\label{eq:error_deriv}
\frac{\partial C}{\partial z_\mathrm{out}(i)} = \sum_{j \in V} \frac{\partial C}{\partial y_\mathrm{out}(j)} \frac{\partial y_\mathrm{out}(j)}{z_\mathrm{out}(i)} = y_\mathrm{out}(i) - t_{i} .
\end{equation}
The further derivation of Eq. (\ref{eq:error_deriv}) requires splitting $\frac{\partial y_\mathrm{out}(j)}{z_\mathrm{out}(i)}$ into cases of $i=j$ and $i \neq j$, i.e. $\frac{\partial y_\mathrm{out}(i)}{\partial z_\mathrm{out}(i)} = y_\mathrm{out}(i)(1-y_\mathrm{out}(i))$ vs. $\frac{\partial y_\mathrm{out}(i)}{\partial z_\mathrm{out}(j)} = -y_\mathrm{out}(i)y_\mathrm{out}(j)$ and is omitted here. For simplicity of presentation, the following equations omit the indices $i,j$.

To back-propagate derivatives from the output layer to the hidden layer, we follow the order $\frac{\partial C}{\partial z_\mathrm{out}}$ $\rightarrow$ $\frac{\partial C}{\partial W_\mathrm{hid-out}}$, $\frac{\partial C}{\partial b_\mathrm{out}}$ $\rightarrow$ $\frac{\partial C}{\partial y_\mathrm{hid}}$ $\rightarrow$ $\frac{\partial C}{\partial Z_\mathrm{hid}}$. Since $Z_\mathrm{out} = W_\mathrm{hid-out}^{T} \cdot y_\mathrm{hid}$, then $\frac{\partial z_\mathrm{out}}{\partial w_\mathrm{hid-out}} = y_\mathrm{hid}$ and $\frac{\partial z_\mathrm{out}}{\partial y_\mathrm{hid}} = w_\mathrm{hid-out}$. In addition, since Eq. (\ref{eq:out_input}), then $\frac{\partial z_\mathrm{out}}{\partial b_\mathrm{out}} = 1$. Thus, 
\begin{equation}
\label{eq:hid_out_gradient}
\frac{\partial C}{\partial w_\mathrm{hid-out}} 
   = \frac{\partial z_\mathrm{out}}{\partial w_\mathrm{hid-out}} \cdot \frac{\partial C}{\partial z_\mathrm{out}}
   = y_\mathrm{hid} \frac{\partial C}{\partial z_\mathrm{out}} ,
\end{equation}
\begin{equation}
\label{eq:out_bias_gradient}
\frac{\partial C}{\partial b_\mathrm{out}} 
   = \frac{\partial C}{\partial z_\mathrm{out}} \cdot \frac{\partial z_\mathrm{out}}{\partial b_\mathrm{out}}
   = \frac{\partial C}{\partial z_\mathrm{out}} ,
\end{equation}
and
\begin{equation}
\label{eq:back_deriv_y1}
\frac{\partial C}{\partial y_\mathrm{hid}} 
  = \sum_{N_\mathrm{out}} \frac{\partial z_\mathrm{out}}{\partial y_\mathrm{hid}} \cdot \frac{\partial C}{\partial z_\mathrm{out}}
  = \sum_{N_\mathrm{out}} w_\mathrm{hid-out} \frac{\partial C}{\partial z_\mathrm{out}} .
\end{equation}
Also, 
\begin{equation}
\label{eq:back_deriv_z1}
\frac{\partial C}{\partial z_\mathrm{hid}} = \frac{\partial C}{\partial y_\mathrm{hid}} \cdot 
  \frac{d y_\mathrm{hid}}{d z_\mathrm{hid}} ,
\end{equation}
where $\frac{d y_\mathrm{hid}}{d z_\mathrm{hid}} = y_\mathrm{hid}(1-y_\mathrm{hid})$, derived using Eq. (\ref{eq:hid_output}). 

To back propagate derivatives from the hidden layer to the embedding layer, the derivations of $\frac{\partial C}{\partial w_\mathrm{emb-hid}}$, $\frac{\partial C}{\partial b_\mathrm{hid}}$ and $\frac{\partial C}{\partial y_\mathrm{emb}}$ are very similar to Eq. (\ref{eq:hid_out_gradient}) through Eq. (\ref{eq:back_deriv_y1}), so that
\begin{equation}
\label{eq:emb_hid_gradient}
\frac{\partial C}{\partial w_\mathrm{emb-hid}} 
   = \frac{\partial z_\mathrm{hid}}{\partial w_\mathrm{emb-hid}} \cdot \frac{\partial C}{\partial z_\mathrm{hid}}
   = y_\mathrm{emb} \frac{\partial C}{\partial z_\mathrm{hid}} ,
\end{equation}
\begin{equation}
\label{eq:hid_bias_gradient}
\frac{\partial C}{\partial b_\mathrm{hid}} 
   = \frac{\partial C}{\partial z_\mathrm{hid}} \cdot \frac{\partial z_\mathrm{hid}}{\partial b_\mathrm{hid}}
   = \frac{\partial C}{\partial z_\mathrm{hid}} ,
\end{equation}
and
\begin{equation}
\label{eq:back_deriv_y2}
\frac{\partial C}{\partial y_\mathrm{emb}} 
  = \sum_{N_\mathrm{hid}} \frac{\partial z_\mathrm{hid}}{\partial y_\mathrm{emb}} \cdot \frac{\partial C}{\partial z_\mathrm{hid}}
  = \sum_{N_{\mathrm{hid}}} w_\mathrm{emb-hid} \frac{\partial C}{\partial z_\mathrm{hid}} .
\end{equation}
However, since the embedding layer is linear rather than sigmoid, then $\frac{d y_\mathrm{emb}}{d z_\mathrm{emb}} = 1$. Thus, 
\begin{equation}
\label{eq:back_deriv_z2}
\frac{\partial C}{\partial z_\mathrm{emb}} = \frac{\partial C}{\partial y_\mathrm{emb}} \cdot 
  \frac{d y_\mathrm{emb}}{d z_\mathrm{emb}} = \frac{\partial C}{\partial y_\mathrm{emb}} .
\end{equation}

In the back propagation from the embedding layer to the word layer, since $W_\mathrm{word-emb}$ is shared among all words, to obtain $\frac{\partial C}{\partial W_\mathrm{word-emb}}$, $\frac{\partial C}{\partial z_\mathrm{emb}}$ needs to be segmented into $\frac{\partial C}{\partial z_\mathrm{emb}(i)}$, such as $[ (\frac{\partial C}{\partial z_\mathrm{emb}(1)} )^T \cdots (\frac{\partial C}{\partial z_\mathrm{emb}(i)} )^T \cdots \frac{\partial C}{\partial z_\mathrm{emb}(N)} )^T ]^{T}$, where $i \in N,i \neq t$ is the index for each input word. 
From Eq. (\ref{eq:emb_input}), $\frac{\partial z_\mathrm{emb}}{\partial w_{word-emb}} = x_{i}$, and then
\begin{equation}
\label{eq:word_emb_gradient}
\frac{\partial C}{\partial w_\mathrm{word-emb}} 
  = \sum_{i \in N, i \neq t} x_{i} \cdot \frac{\partial C}{\partial z_\mathrm{emb}(i)} .
\end{equation}

\subsection{Weight and Bias Update}
After each iteration of forward-backward propagation, the weights and biases are updated to reduce cost. Denote $W$ as a general form of the weight matrices $W_\mathrm{word-emb}$, $W_\mathrm{emb-hid}$ and $W_\mathrm{hid-out}$, and $\Delta$ as an averaged version of the weight gradient,  which carries information from previous iterations and is initialized with zeros, the weights are updated with:
\begin{equation}
\label{eq:weight_update}
 \left\{ \begin{array}{l}
 \Delta_{i+1} = \alpha \Delta_{i} + \frac{\partial C}{\partial W_{i}} \\
 W_{i+1} = W_{i} - \epsilon \Delta_{i+1} 
 \end{array} \right .
\end{equation}
where $\alpha$ is the momentum which determines the percentage of weight gradients carried from the previous iteration, and $\epsilon$ is the learning rate which determine the step size to update weights towards the direction of descent. The biases are updated similarly by just replacing $W$ with $b$ in Eq. (\ref{eq:weight_update}).

\subsection{Summary of NNLM}
In the NNLM training, the whole training dataset is segmented into mini-batches with batch size $M$. The neural network in terms of weights and biases gets updated through each iteration of mini-batch training. The gradient $\frac{\partial C}{\partial W_{i}}$ in Eq. (\ref{eq:weight_update}) should be normalized by $M$. One cycle of feeding all data is called an epoch, and given appropriate training parameters such as learning rate $\epsilon$ and momentum $\alpha$, it normally requires $10$ to $20$ epochs to get a well-trained network. 

Next we present a procedure for training the NNLM. It includes all the key components described before, has the flexibility to change the training parameters through different epochs, and includes an early termination criterion.
\begin{enumerate}[1.]\itemsep0pt
\item Set up general parameters such as the mini-batch size $M$, the number of epochs and model parameters such as the word context size $N$, the target word position $t$, the number of nodes in each layer, etc.;
\item Split the training data into mini-batches;
\item Initialize networks, such as weights and biases;
\item For each epoch:
  \begin{enumerate}[a.]\itemsep0pt
  \item Set up parameters for current epoch, such as the learning rate $\epsilon$, the momentum $\alpha$, etc.;
  \item For each iteration of mini-batch training:
  		\begin{enumerate}[i.]\itemsep0pt
  		\item Compute weight and bias gradients through forward-backward propagation;
  		\item Update weights and biases with current $\epsilon$ and $\alpha$.
  		\end{enumerate}
  \item Check the cost reduction of the validation set, and terminate the training early, if it goes up.	
  \end{enumerate}
\end{enumerate}

\section{\uppercase{Implementation and Results}}
\label{sec:results}

\noindent This section covers the implementation details of the authorship attribution system as a $N$-way classification problem using NNLM. The results are compared with baseline N-gram language models trained using the SRILM toolkit \cite{stolcke2002srilm}. 

\subsection{NNLM Implementation and Optimization}
\label{subsec: implementation}

The  database for each of the 16 courses is randomly split into training, validation and test sets with ratio 8:1:1. To compensate for the model variation due to the limited data size, the segmentation is performed 10 times with different randomization seeds, so the mean and variation of performance can be measured.  

For each course in this project, we trained a different 4-gram NNLM, i.e. context size $N=4$, to predict the 4th word using the 3 preceding words. These models share the same general parameters, such as a) the number of epochs (15), b) the epoch in which the learning rate decay starts (10), c) the learning rate decay factor (0.9). However, the other model parameters are searched and optimized within certain ranges using a multi-resolutional optimization scheme, with a) the dimension of embedding space $N_\mathrm{emb}$ ($25 \sim 200$), b) the nodes of the hidden layer $N_\mathrm{hid}$ ($100 \sim 800$), c) the learning rate $\epsilon$ ($0.05 \sim 0.3$), d) the momentum $\alpha$ ($0.8 \sim 0.99$), and e) mini-batch size $M$ ($100 \sim 400$). This optimization process is time consuming  but worthwhile, since each course has a unique profile, in terms of vocabulary size, word distribution, database size, etc., so a model adapted to its profile can perform better in later classification.  


\subsection{Classification with Perplexity Measurement}
\label{subsec: perplexity}

Statistical language models provide a tool to compute the probability of the target word $\mathcal{W}_{t}$ given $N-1$ context words $\mathcal{W}_{1}, \mathcal{W}_{2}, \ldots, \mathcal{W}_{i}, \ldots, \mathcal{W}_{N}$, $i \in N, i \neq t$. Normally, the target word is the $N$th word and the context words are the preceding $N-1$ words. Denote $\mathcal{W}_{1}^{n}$ as a word sequence $(\mathcal{W}_{1}, \mathcal{W}_{2}, \ldots, \mathcal{W}_{n})$. Using the chain rule of probability, the probability of sequence $\mathcal{W}_{1}^{n}$ can be formulated as
%
%
\begin{eqnarray}
\begin{aligned}
\label{eq:seq_prob}
P(\mathcal{W}_{1}^{n}) & = P(\mathcal{W}_{1})P(\mathcal{W}_{2}|\mathcal{W}_{1}) \ldots P(\mathcal{W}_{n}|\mathcal{W}_{1}^{n-1}) \\ 
& = \prod_{k=1}^{n}P(\mathcal{W}_{k}|\mathcal{W}_{1}^{k-1}). 
\end{aligned}
\end{eqnarray}

Using a Markov chain, which approximates the probability of a word sequence with arbitrary length $n$ to the probability of a sequence with the closest $N$ words, the shortened probabilities can be provided by the LM with context size $N$, i.e. N-gram language model. Eq. (\ref{eq:seq_prob}) can then be simplified to
\begin{equation}
\label{eq:seq_prob_ngram}
P(\mathcal{W}_{1}^{n}) \approx P(\mathcal{W}_{n-N+1}^{n}) = \prod_{k=1}^{n}P(\mathcal{W}_{k}|\mathcal{W}_{k-N+1}^{k-1})
\end{equation}

Perplexity is an intrinsic measurement to evaluate the fitness of the LM to the test word sequence $\mathcal{W}_{1}^{N}$, which is defined as
\begin{equation}
\label{eq:ppl1}
PP(\mathcal{W}_{1}^{n}) = P(\mathcal{W}_{1}^{n})^{-\frac{1}{n}}
\end{equation}
In practical use, it normally converts the probability multiplication to the summation of log probabilities. Therefore, using Eq. (\ref{eq:seq_prob_ngram}), Eq. (\ref{eq:ppl1}) can be reformulated as
\begin{eqnarray}
\begin{aligned}
\label{eq:ppl2}
PP(\mathcal{W}_{1}^{n}) & \approx  \left( \prod_{k=1}^{n} P(\mathcal{W}_{k}|\mathcal{W}_{k-N+1}^{k-1}) \right) ^ {-\frac{1}{n}}  \\
  & = 10 ^ {-\dfrac{\sum_{k=1}^{n}\log_{10}P(\mathcal{W}_{k}|\mathcal{W}_{k-N+1}^{k-1})}{n}}
\end{aligned}
\end{eqnarray}

In this project, the classification is performed by measuring the perplexity of the test word sequences in terms of sentences, using the trained NNLM of each course. Denote $\mathcal{C}$ as the candidate courses/instructors and $\mathcal{C}^{*}$ as the selected one from the classifier. $\mathcal{C}^{*}$ can then be expressed as 
\begin{equation}
\label{eq: classifier}
\mathcal{C}^{*} = \arg\max_{\mathcal{C}} PP(\mathcal{W}_{1}^{n}|\mathrm{LM}_\mathcal{C})
\end{equation}
The classification performance with NNLM is also compared with baselines from an SRI N-gram back-off model with Kneser-Ney Smoothing. The perplexities are computed without insertions of start-of-sentence and end-of-sentence tokens in both SRILM and NNLM. To evaluate the LM fitness with different training methods, Table \ref{tab:ppl_comp} lists the training-to-test perplexities for each of the 16 courses, averaged from 10 different database segmentations.
\begin{table}[!htb]
\centering
\caption{Perplexity comparison with different LM training methods}
\label{tab:ppl_comp}
\setlength{\tabcolsep}{2pt}
\scriptsize
\renewcommand{\arraystretch}{1}
\begin{tabular}{@{} *{6}{l} @{}} \toprule%
\multirow{2}*{ID} & \multicolumn{4}{c}{SRI N-gram} & NNLM \\
  & unigram & bigram & trigram & 4-gram & 4-gram \\ \midrule
1 & 251.7 $\pm$ 3.5 & 84.7 $\pm$ 2.5 & 75.3 $\pm$ 2.4 & 75.0 $\pm$ 2.3 & 71.1 $\pm$ 1.3 \\
2 & 301.9 $\pm$ 3.2 & 84.4 $\pm$ 1.9 & 69.7 $\pm$ 1.8 & 68.5 $\pm$ 1.8 & 63.9 $\pm$ 1.8 \\
3 & 186.2 $\pm$ 2.1 & 49.8 $\pm$ 1.5 & 43.6 $\pm$ 1.8 & 43.2 $\pm$ 1.8 & 40.2 $\pm$ 1.7 \\
4 & 283.2 $\pm$ 5.3 & 82.9 $\pm$ 2.2 & 74.1 $\pm$ 2.0 & 74.1 $\pm$ 2.0 & 77.2 $\pm$ 2.1 \\
5 & 255.7 $\pm$ 3.2 & 75.2 $\pm$ 1.2 & 65.8	$\pm$ 1.5 & 65.4 $\pm$ 1.4 & 62.7 $\pm$ 1.7 \\
6 & 273.4 $\pm$ 3.9 & 85.3 $\pm$ 1.8 & 72.9 $\pm$ 1.8 & 71.8 $\pm$ 1.8 & 72.8 $\pm$ 1.3 \\
7 & 300.9 $\pm$ 7.8 & 122.2 $\pm$ 3.4 & 114.0 $\pm$ 3.0 & 114.1 $\pm$ 3.0 & 110.1 $\pm$ 2.8 \\
8 & 209.6 $\pm$ 7.1 & 57.8 $\pm$ 2.5 & 47.0 $\pm$ 2.2 & 45.9 $\pm$ 2.2 & 48.0 $\pm$ 2.1 \\
9 & 255.9 $\pm$ 4.0 & 69.2 $\pm$ 2.6 & 57.6 $\pm$ 2.5 & 57.1 $\pm$ 2.4 & 53.2 $\pm$ 1.8 \\
10 & 243.3 $\pm$ 3.0 & 83.5 $\pm$ 1.7 & 74.1 $\pm$ 1.7 & 73.7 $\pm$ 1.7 & 72.2 $\pm$ 1.5 \\
11 & 272.4 $\pm$ 4.8 & 93.1 $\pm$ 2.1 & 84.7 $\pm$ 1.9 & 84.7 $\pm$ 1.9 & 80.5 $\pm$ 1.7 \\
12 & 247.1 $\pm$ 10.7 & 78.2 $\pm$ 7.8 & 68.6 $\pm$ 8.2 & 67.2 $\pm$ 8.5 & 70.5 $\pm$ 12.2 \\
13 & 237.3 $\pm$ 3.4 & 61.9	$\pm$ 1.4 & 50.4 $\pm$ 1.1 & 49.7 $\pm$ 1.1 & 48.3 $\pm$ 1.5 \\
14 & 301.0 $\pm$ 6.5 & 91.8 $\pm$ 3.0 & 83.0 $\pm$ 3.1 & 82.5 $\pm$ 3.1 & 79.4 $\pm$ 2.0 \\
15 & 308.4 $\pm$ 4.1 & 88.4 $\pm$ 1.0 & 69.3 $\pm$ 0.9 & 67.5 $\pm$ 0.9 & 65.5 $\pm$ 1.5 \\
16 & 224.1 $\pm$ 4.4 & 74.5 $\pm$ 2.2 & 64.8 $\pm$ 2.1 & 64.6 $\pm$ 2.2 & 61.8 $\pm$ 1.8 \\ \midrule
Avg. & 259.5 $\pm$ 4.8 & 80.2 $\pm$ 2.4 & 69.7 $\pm$ 2.4 & 69.0 $\pm$ 2.4 & 67.3 $\pm$ 2.4 \\ \bottomrule
\end{tabular}
\end{table}
Each line in Table \ref{tab:ppl_comp} shows the mean perplexities with standard deviation for the SRI N-gram methods with $N$ from 1 to 4, plus the NNLM 4-gram method. It illustrates that among the SRI N-gram methods, 4-gram is slightly better than the tri-gram, and for the 4-gram NNLM method, it achieves even lower perplexities on average. 

\subsection{Classification Accuracy and Confusion Matrix}

\begin{figure}[!htb]
\centering
\includegraphics[width=0.48\textwidth]{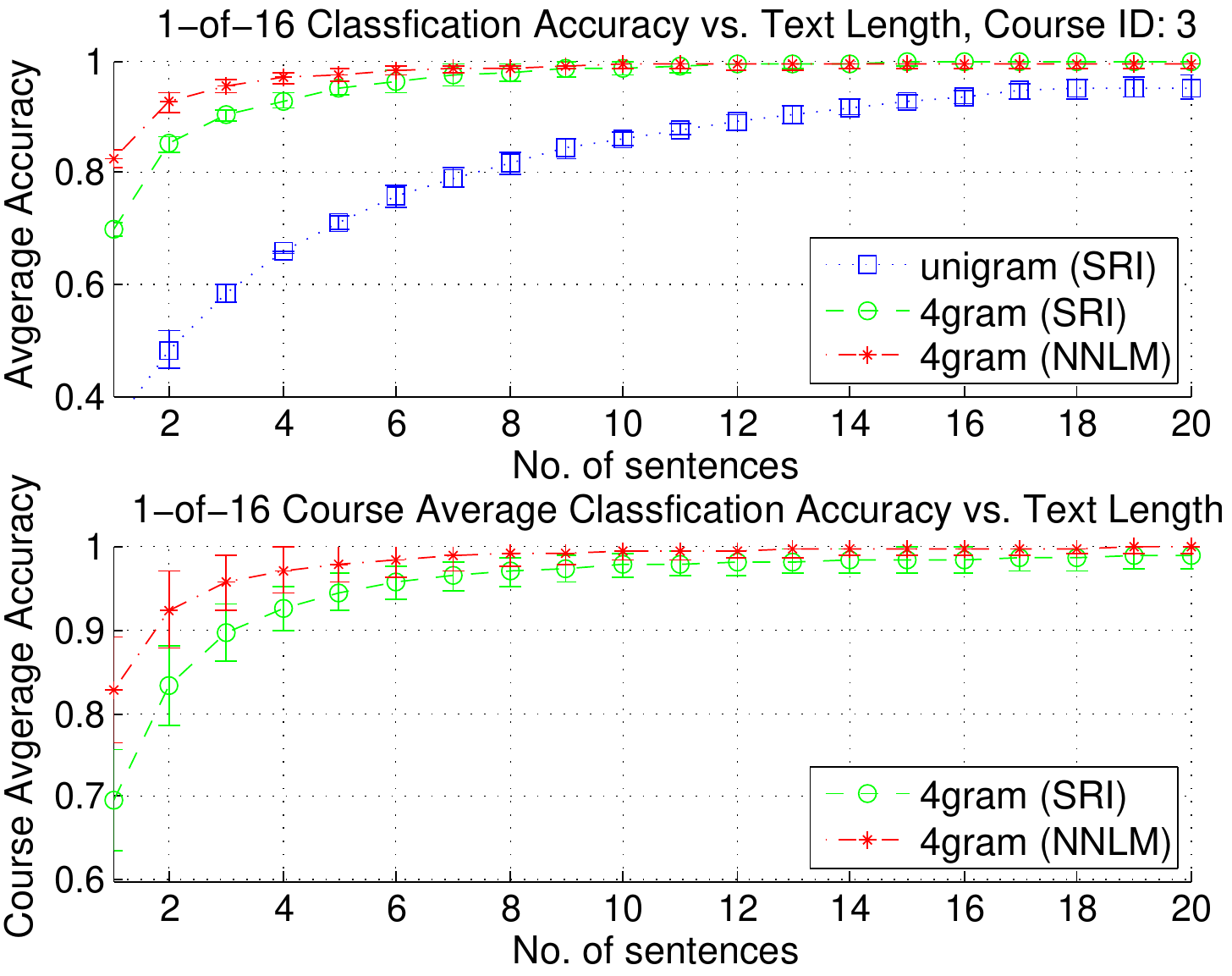}
\caption{Individual (a) and mean (b) accuracies vs. text length in terms of the number of sentences}
\label{fig:acc_sent}
\end{figure}

To test the classification accuracy for a  particular course instructor, the sentence-wise perplexity is computed with the trained NNLMs from different classes. The sentences are randomly selected from the test set. Figure \ref{fig:acc_sent}(a) shows graphically the accuracy vs. number of sentences for a particular course with ID 3. The accuracies are obtained from 3 different methods, SRI uniqram, 4-gram and NNLM 4-gram. The number of randomly selected sentences is in the range of 1 to 20, and for each particular number of sentences, 100 trials were performed and the mean accuracies with standard deviations are shown in the figure. As mentioned earlier in Sec. \ref{sec:data}, courses with ID 7 and 16 were taught by the same instructor, so these two courses are excluded and 14 courses/instructors are used to compute their 16-way classification accuracies. Figure \ref{fig:acc_sent}(b) demonstrates the mean accuracy over these 14 courses. SRI 4-gram and NNLM 4-gram achieve similar accuracy and variation. However, the NNLM 4-gram is slightly more accurate than the SRI 4-gram.     

\begin{figure}[htb]
\centering
\includegraphics[width=0.49\textwidth]{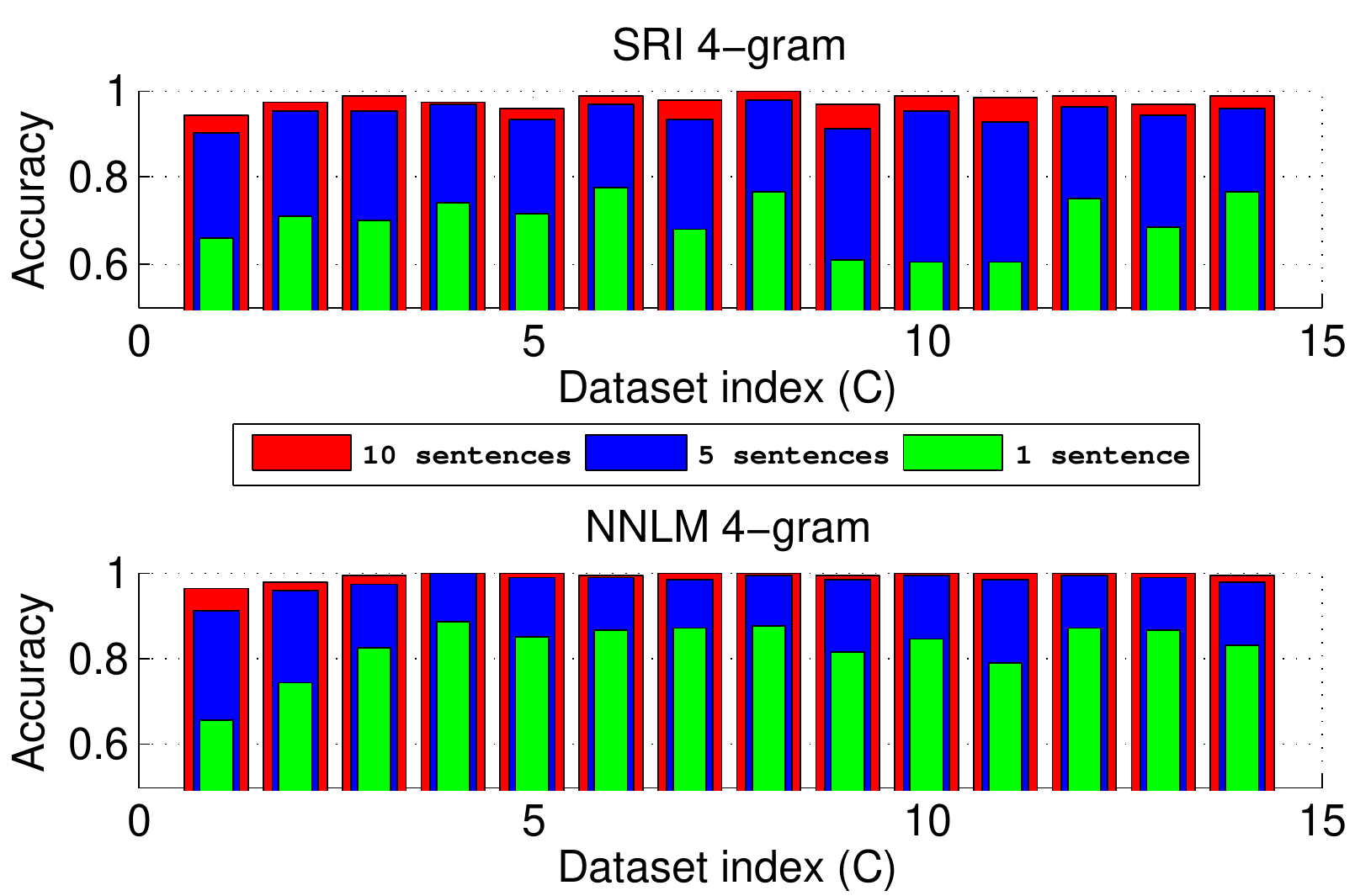}
\caption{Accuracies at 3 stages differed by text length for each of the 14 courses. The 2 courses taught by the same instructor are excluded}
\label{fig:acc_stages}
\end{figure}

Figure \ref{fig:acc_stages} again compares the accuracies from these two models. It provides the accuracies of 3 difficulty stages, given 1, 5, or 10 test sentences. Both LMs perform differently along all course/instructor datasets. However, NNLM 4-gram is on average slightly better than SRI 4-gram, especially when the number of sentences is less. 

\begin{figure}[htb]
\begin{minipage}{0.95 \linewidth}
  \centering
  \centerline{\hspace{0em}\includegraphics[height=4.8cm, width=7.5cm]{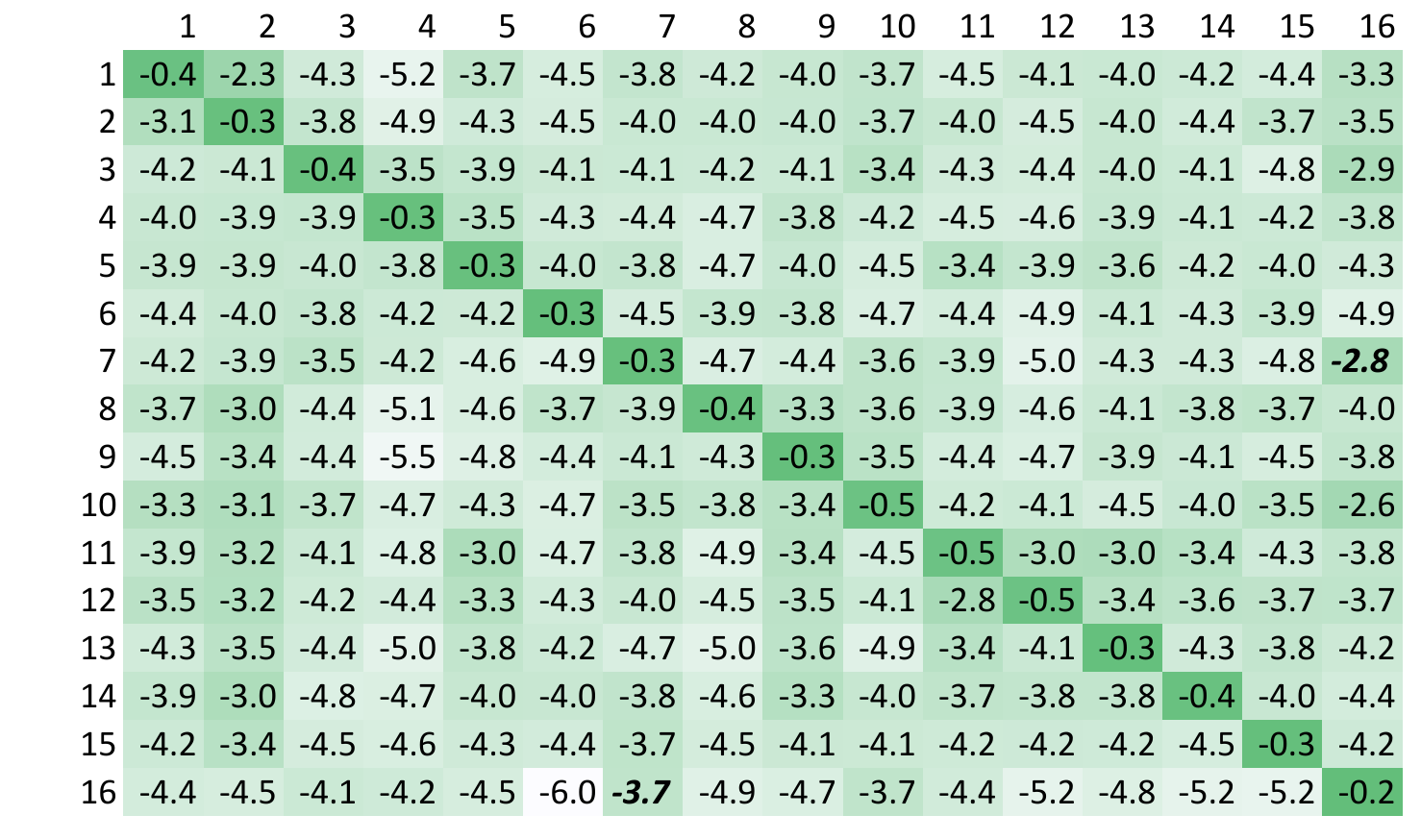}}
  \centerline {\hspace{0em} \small {(a)}}\medskip
\end{minipage}
\hfill
\begin{minipage}{0.95 \linewidth}
  \centering
  \centerline{\hspace{0em}\includegraphics[height=4.8cm, width=7.5cm]{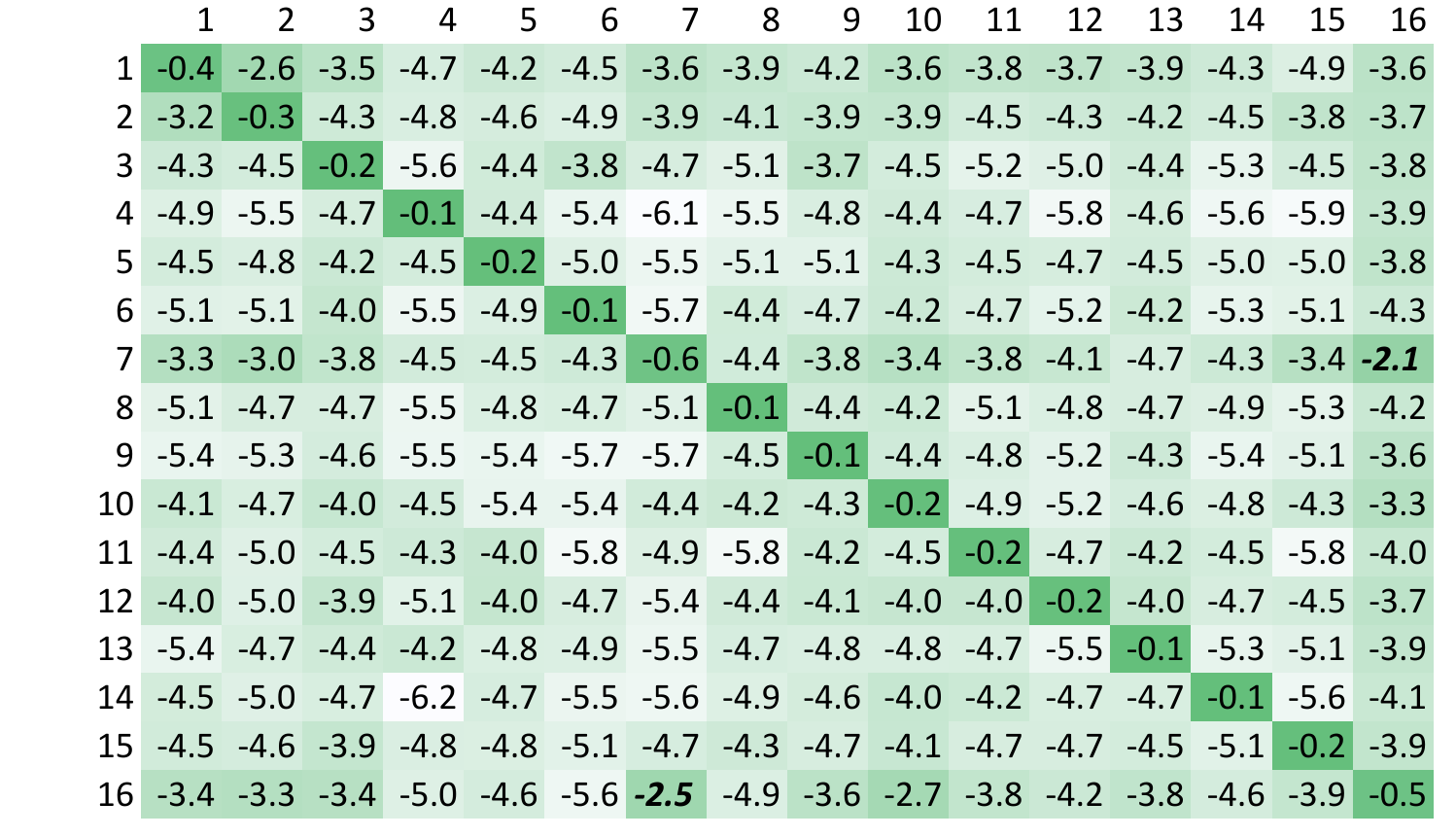}}
  \centerline {\hspace{0em} \small {(b)}}\medskip
\end{minipage}
\caption{\small Course/instructor confusion matrices ($16 \times 16$) for SRI 4-gram (a) and NNLM 4-gram (b).
\label{fig:confmtx}}  
\end{figure}

Besides classification accuracy, the confusion between different course/instructors is also investigated. Figure \ref{fig:confmtx} shows the confusion matrices for all 16 courses/instructors, computed with only one randomly picked test sentence for both methods. The probabilities are all in log scale for better visualization. The confusion value for the $i$th row and $j$th column is the log probability of assigning the $i$th course/instructor as the $j$th one. Since course 7 and 16 were taught by the same instructor, it is not surprising that the values for $(7,16)$ and $(16,7)$ are larger than the others in the same row. In addition, instructors who taught the courses in the same field, such as courses $1, 2$ (Algorithm) and courses $11, 12, 13$ (NLP) are more likely to be confused with each other. So the topic of the text does play a big role in authorship attribution. Since the NNLM 4-gram assigns higher values than the SRI the 4-gram for $(7,16)$ and $(16,7)$, it is more biased towards the author rather than the content in that sense. 

\section{\uppercase{Conclusion and Future Work}}
\label{sec:conclusion}

\noindent This paper investigates authorship attribution using NNLM. The experimental setup for NNLM is detailed with mathematical elaboration. The results in terms of LM fitness in perplexity, classification accuracies, and confusion scores are promising, compared with the baseline N-gram methods. The performance is very competitive to the state-of-the-art, in terms of classification accuracy and testing sensitivity, i.e. the length of test text used in order to achieve confident results. From the previous work listed in Sec. \ref{sec:introduction}, the best reported results to date achieved either $95$\% accuracy on a similar author pool size, or $50\% \sim 60\%$ with 100+ authors and limited training date per author. As it is shown in Figure \ref{fig:acc_stages}, our work achieves nearly perfect accuracies if more than 10 test sentences are given.

However, since both the SRI baseline and NNLM methods achieves nearly perfect accuracies with only limited test data, the current database may not be sufficiently large and challenging, probably due to the consistency between the training and the test sets and the contribution from the topic distinction. In the future, the algorithm should be tested using datasets with larger author set sizes and greater styling similarities.

Since purely topic-neutral text data may not even exist \cite{luyckx2011scalability}, developing general author LMs with mixed-topic data, and then adapting them to particular topics may also be desirable. It could be particularly helpful when the topics of text data is available. To compensate the relatively small size of the training set, LMs may also be trained with a group of authors and then adapt to the individuals.

Because the NNLM assigns a unique representation for a single word, it is difficult to model words with multiple meanings \cite{mnih2010learning}. Thus, combining the NNLM and N-gram models might be beneficial. The recurrent NNLM, which captures more context size than the current feed-forward model \cite{mikolov2010recurrent}, may also be worth exploring. 

\section*{\uppercase{Acknowledgements}}

\noindent The authors would like to thank Coursera Incorporation for providing the course transcript datasets for the research in this paper.

\vfill
\bibliographystyle{apalike}
{\small
\bibliography{paper2}}

\vfill
\end{document}